\documentclass[conference,10pt]{IEEEtran}

\usepackage{amsmath}
\usepackage{amssymb}
\usepackage{subcaption}
\usepackage{graphicx}
\usepackage{epstopdf}
\DeclareMathOperator*{\argmin}{arg\,min}

\usepackage{array}
\usepackage{amsthm}
\usepackage{algpseudocode}
\usepackage{algorithm}
\DeclareMathAlphabet\mathbfcal{OMS}{cmsy}{b}{n}

\IEEEoverridecommandlockouts

\begin{document}

\title{Completing a joint PMF from projections: a low-rank coupled tensor factorization approach}

\author{Nikos Kargas,~\IEEEmembership{Student~Member,~IEEE}, Nicholas D. Sidiropoulos,~\IEEEmembership{Fellow,~IEEE}
\thanks{N. Kargas and N.D. Sidiropoulos are with the ECE Department, University of Minnesota, Minneapolis, USA; e-mail: {\tt (karga005,nikos)@umn.edu}. Supported in part by NSF IIS-1247632, IIS-1447788.}
}

\maketitle
\begin{abstract}
There has recently been considerable interest in completing a low-rank matrix or tensor given only a small fraction (or few linear combinations) of its entries. Related approaches have found considerable success in the area of recommender systems, under machine learning. From a statistical estimation point of view, the gold standard is to have access to the joint probability distribution of all pertinent random variables, from which any desired optimal estimator can be readily derived. In practice high-dimensional joint distributions are very hard to estimate, and only estimates of low-dimensional projections may be available. We show that it is possible to identify higher-order joint PMFs from lower-order marginalized PMFs using coupled low-rank tensor factorization. Our approach features guaranteed identifiability when the full joint PMF is of low-enough rank, and effective approximation otherwise. We provide an algorithmic approach to compute the sought factors, and illustrate the merits of our approach using rating prediction as an example.
\end{abstract}
\IEEEpeerreviewmaketitle

\section{Introduction}
Consider the setting where we are given a partially observed dataset of $M$ discrete samples $\left( x_1^{(i)},x_2^{(i)},\ldots,x_N^{(i)} \right)$, $i=1,\ldots,M,$ and we are interested in predicting the missing entries. This scenario often arises in recommender systems where we are interested in predicting user preferences, pertaining to news, movies or music, based on a user's history as well as the history of other users.

Among the various approaches used in recommender systems, data completion using matrix and (more recently) tensor factorization is one of the most pervasive. The premise of factorization-based recommendation approaches is that the data approximately follow a low-rank model, i.e., there are few basic types of customers (and movies, songs, or news items), and every customer (movie, song, story) is a linear combination of the respective types. Thus a low-rank model is appropriate for the data, and can be used for completion.

In this paper, we propose a fundamentally different approach. From a statistical inference point of view, having access to the joint distribution of all variables of interest is the `gold standard'. Given the joint Probability Mass Function  (PMF) of a set of discrete random variables, it is possible to compute any marginal or conditional probability for subsets of these variables, and use it to solve regression or classification problems. For example, one may be interested in finding the Maximum A Posteriori (MAP) estimate of an unobserved entry, or its conditional expectation given a number of observed variables. In practice however, it is often not possible to learn a joint PMF of all random variables without making restrictive assumptions, due to computational or statistical reasons; the number of free parameters grows exponentially in the number of variables. In addition, when the dataset is incomplete an imputation mechanism is needed.

In this work, we propose modeling a joint PMF of a set of random variables using a low-rank non-negative tensor (multi-way array) factorization model. In effect, we propose using a low-rank model of the joint PMF, as opposed to using a low-rank model of the raw data. Tensor factorization techniques are widely used in numerous diverse fields such as signal processing, computer vision, chemistry and more recently in machine learning and data mining~\cite{SiDeFu2016}. Canonical Polyadic Decomposition (CPD) also known as PARAllel FACtor analysis (PARAFAC)~\cite{CaCha1970,HaLu1994} and the Tucker decomposition~\cite{Tucker1966} are the two most widely used factorization models. In this work, we focus on the CPD model which is known to be unique under mild conditions.

Any joint PMF of size $I_1 \times \cdots \times I_N$ can be regarded as a non-negative CPD model of non-negative rank 
$\leq \underset{k}{\min}(\prod_{\substack{n=1 \\ n \neq k}}^N I_n)$ -- this is easy to see, using the same argument as for real-valued CPD in \cite{SiDeFu2016}, and the trivial factorization ${\bf A}={\bf A} {\bf I}$. Symmetric CPD has been considered for the related (but different) problem of modeling co-occurrence data \cite{ShaHa2005,AnGeHsu2014}. The most relevant prior work is \cite{DuXi2009} (and \cite{ChiZhu2008}), where CPD (resp. Tucker) was used to model the joint PMF of multivariate categorical data, assuming access to the full joint PMF.

We do not assume access to the full joint PMF. The reason is that, when dealing with many random variables (large $N$), the probability of encountering any particular realization decays very fast (usually exponentially) as a function of $N$. That makes joint PMF estimation, even from complete samples, essentially intractable. One needs very long data records (very high $M$) for reliable estimates of the joint PMF values. In this paper, we show how we can utilize information regarding lower-order marginals of subsets of the random variables to provably infer the full joint PMF. Estimates of lower-order marginals are easier to compute even in the case of missing data, and a low-rank CPD model has the advantage of reducing the dimension of the parameter space, which becomes linear in the number of variables. We formulate the problem as a coupled tensor factorization problem and derive an algorithmic approach to solve it. We illustrate the method using synthetic data and give a motivating example using real data for rating prediction.

\subsection{Notation} 
$\mathbf{x}$ denotes a vector, $\mathbf{X}$ denotes a matrix and $\underline{\mathbf{X}}$ denotes a tensor. 
The outer product of $N$ vectors is a $N$-way tensor with elements $(\mathbf{a}_1 \circ \mathbf{a}_2 \cdots \circ \mathbf{a}_N)(i_1,i_2,\ldots,i_N) = \mathbf{a}_1(i_1)\mathbf{a}_2(i_2)\cdots \mathbf{a}_N(i_N)$. The Khatri-Rao (columnwise Kronecker) product of two matrices $\mathbf{A} \in \mathbb{R}^{I \times F}$ and $\mathbf{B} \in \mathbb{R}^{J \times F}$ is $\mathbf{A} \odot \mathbf{B}  \in \mathbb{R}^{IJ \times F} $. The Hadamard  (elementwise) product of commensurate matrices is denoted $\mathbf{A} \circledast \mathbf{B}$. $\text{vec}(\mathbf{X})$ is the vector obtained by vertically stacking the columns of matrix $\mathbf{X}$. $\mathcal{D}(\mathbf{x}) \in \mathbb{R}^{I \times I}$ denotes the diagonal matrix with the elements of vector $\mathbf{x}  \in \mathbb{R}^{I}$ on its diagonal.

\section{Tensor Decomposition Preliminaries}
\begin{figure}
\centering
\includegraphics[width= 0.45 \textwidth]{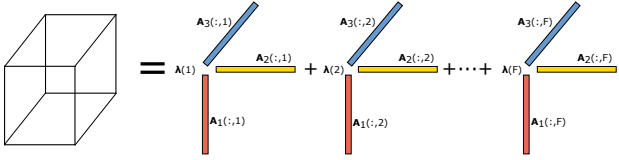}
\caption{CPD model.}
\label{fig:CPmodel}
\end{figure}

$N$-way tensor $\underline{\mathbf{X}} \in \mathbb{R}^{I_1 \times I_2 \times \cdots \times I_N}$ admits a CPD of rank $F$ if it can be decomposed as a sum of $F$ rank-1 tensors
\begin{multline}
\underline{\mathbf{X}} = \sum_{f=1}^F \boldsymbol{\lambda}(f) \mathbf{A}_1(:,f)\circ \mathbf{A}_2(:,f) \circ \cdots  \circ \mathbf{A}_N(:,f),
\label{eq:CP_model}
\end{multline}
where $\boldsymbol{\lambda} \in \mathbb{R}^F, \mathbf{A}_n \in \mathbb{R}^{I_n \times F}, n=1,2,\ldots,N$ and $F$ is the smallest number for which such a decomposition exists (Fig.~\ref{fig:CPmodel}). We then write $\underline{\mathbf{X}} = [\![ \boldsymbol{\lambda},\mathbf{A}_1,\ldots,\mathbf{A}_N ]\!]$, with elements
\begin{multline}
\underline{\mathbf{X}}(i_1,\ldots,i_N) = \\ \sum_{f=1}^F \boldsymbol{\lambda}(f)  \mathbf{A}_1(i_1,f) \mathbf{A}_2(i_2,f) \cdots  \mathbf{A}_N(i_N,f).
\label{eq:individ}
\end{multline}
We denote the mode-$n$ matrix unfolding of $\underline{\mathbf{X}}$ as the matrix ${\mathbf{X}}^{(n)}$ of size $\prod_{ \substack{k=1 \\ k\neq n}}^N I_k \times I_n$. We have that $\underline{\mathbf{X}}(i_1,i_2,\ldots,i_N) = {\mathbf{X}}^{(n)}(j,i_n)$, where
\begin{equation}
j = 1 + \sum_{ \substack{k=1 \\ k \neq n} }^N (i_k-1) J_k \; \text{with} \; J_k = \prod_{  \substack{m=1 \\ m \neq n}}^{k-1} I_m.
\end{equation}
The mode-$n$ matrix unfolding can be expressed as
\begin{equation}
{\mathbf{X}}^{(n)} =  \left( \underset{j\neq n}{ \underset{j=1}{ \overset{N}{\odot}}} \mathbf{A}_j \right) \mathcal{D}(\boldsymbol{\lambda}) \mathbf{A}_n^T,
\end{equation}
where
\begin{equation}
\underset{j\neq n}{ \underset{j=1}{ \overset{N}{\odot}}} \mathbf{A}_j = \mathbf{A}_N \odot \cdots \odot \mathbf{A}_{n+1} \odot \mathbf{A}_{n-1} \odot \cdots \odot \mathbf{A}_1.
\end{equation}
We can also express a tensor in a vectorized form. $\underline{\mathbf{X}}(i_1,i_2,\ldots,i_N) = {\mathbf{x}}(j)$, where
\begin{equation}
j = 1 + \sum_{k=1 }^N (i_k-1) J_k \; \text{with} \; J_k = \prod_{m=1}^{k-1} I_m.
\end{equation}
The vectorized form of a tensor can be expressed as
\begin{equation}
\text{vec}(\underline{\mathbf{X}})=  \left( { \underset{j=1}{ \overset{N}{\odot}}} \mathbf{A}_j \right) \boldsymbol{\lambda}.
\end{equation}
\begin{figure}[!t]
\centering
\includegraphics[width=0.2 \textwidth ]{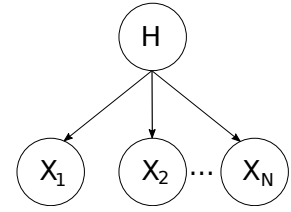}
\caption{Naive Bayes model.}
\label{fig:naive_model}
\end{figure}
\section{Non-Negative Tensor Factorization and Latent Variable Models}
\label{sec:NTF_latent}
It has been shown that the joint PMF of a finite set of discrete random variables satisfying the naive Bayes hypothesis can be regarded as a non-negative CPD model~\cite{ShaHa2005,LiCo2009}. More specifically, let $\{X_1, X_2,\ldots,X_N\}$ denote $N$ discrete random variables that can take one of $I_1$, $I_2$,$\ldots,I_N$ distinct values respectively. We define a $N$-way tensor $\underline{\mathbf{X}} \in \mathbb{R}^{I_1 \times I_2 \times \cdots \times I_N}$ that models the joint PMF of the $N$ random variables i.e., $\underline{\mathbf{X}}(i_1,i_2,\ldots,i_N) = \mathbb{P}(X_1 = i_1, X_2 = i_2,\ldots,X_N = i_N)$. Suppose that the discrete random variables satisfy the naive Bayes hypothesis, that is, they are conditionally independent given a hidden variable $H$ that can take $F$ distinct values (Fig.~\ref{fig:naive_model}). Then, the joint PMF can be decomposed as
\begin{equation}
\begin{aligned}
\mathbb{P}(i_1,i_2,\ldots,i_N) = \sum_{f=1}^F \mathbb{P}(f) \mathbb{P}(i_1 | f) \cdots \mathbb{P}(i_N| f),
\end{aligned}
\label{eq:pmf_latent_var}
\end{equation}
where $\mathbb{P}(f) := \mathbb{P}(H = f) $ is the prior distribution of the latent variable $H$ and $\mathbb{P}(i_n | f) := \mathbb{P}( X_n = i_n | H = f) $, $n=1,\ldots,N$ are conditional distributions. In fact, every joint PMF can be decomposed as in Equation~\eqref{eq:pmf_latent_var} for $F$ large enough, as we explained in the introduction; the naive Bayes hypothesis is just an interpretation. Notice the similarity of equations~\eqref{eq:individ} and~\eqref{eq:pmf_latent_var}. In the CPD model of the joint PMF, each column of the CPD factor matrices is a conditional PMF; and the vector $\boldsymbol{\lambda}$ contains the prior probabilities of the latent variable $H$.

For the case $N=2$, the model described by Equation~\eqref{eq:pmf_latent_var} is equivalent to Probabilistic Latent Semantic Indexing (PLSI)~\cite{Hofmann1999}, a popular method for document clustering based on dyadic co-occurence data which is known to be closely related to NMF with K-L divergence~\cite{GaGo2005}. In the following, we focus on the general case where $N$ can be larger than two and we are interested in cases where the PMF can be approximated by a low-rank CPD model
\begin{equation}
\begin{aligned}
\mathbb{P}(i_1,i_2,\ldots,i_N) \approx \sum_{f=1}^F \mathbb{P}(f) \mathbb{P}(i_1 | f) \cdots \mathbb{P}(i_N| f),
\end{aligned}
\label{eq:pmf_latent_var_aprox}
\end{equation}
ideally for some $F \ll \underset{k}{\min}(\prod_{\substack{n=1 \\ n \neq k}}^N I_n)$.
\section{Problem Formulation}
In practice it is not always possible to have ``point'' estimates of $P(X_1=i_1, X_2=i_2, \ldots, X_N=i_N)$.  When $N$ is large, a very large number of samples is needed in order to obtain a reliable empirical estimate of the PMF. Furthermore, much (most) of the data may be missing, as in collaborative filtering applications. An alternative approach is  to extract estimates for lower-order marginals of subsets of random variables, which can be viewed as linear measurements (sums over the remaining modes, i.e., lower-dimensional {\em projections}) of the complete tensor, and seek a joint PMF that is consistent with this information.

For brevity, we focus on the case where we have estimates of marginal distributions corresponding to every possible combination of triples of random variables, i.e., we are given estimates $\hat{\mathbb{P}}(X_i,X_j,X_k)$, $i,j,k \in \{1,\ldots,N\} $, $i\neq j, i \neq k, j\neq k$ which we put in a tensor $\underline{\mathbf{X}}_{ijk}$
\begin{equation}
\underline{\mathbf{X}}_{ijk}(i',j',k') = \hat{\mathbb{P}}(X_i = i' ,X_j = j' ,X_k = k').
\end{equation}
The method can be easily generalized to any type of lower-order marginals. Under the assumption of a low-rank CPD model as in ~\eqref{eq:pmf_latent_var_aprox}, every marginal distribution of three random variables can be decomposed as follows
\begin{equation}
\mathbb{P}(i',j',k') = \sum_{f=1}^{F}   \mathbb{P}(f) \mathbb{P}(i' | f) \mathbb{P}(j' | f) \mathbb{P}(k' | f).
\end{equation}
This is a direct consequence of the law of total probability. Marginalizing with respect to the $n$-th random variable we have that
\begin{multline}
\sum_{i_n=1}^{I_n}\mathbb{P}(i_1,i_2,\ldots,i_N) =  \sum_{f=1}^F \sum_{i_n=1}^{I_n} \mathbb{P}(f) \mathbb{P}(i_1 | f) \cdots \mathbb{P}(i_N| f) \\
= \sum_{f=1}^F \mathbb{P}(f) \mathbb{P}(i_1 | f) \cdots \mathbb{P}(i_{n-1}| f) \mathbb{P}(i_{n+1}| f) \cdots \mathbb{P}(i_N| f),
\end{multline}
since $\sum_{i_n=1}^{I_n}  \mathbb{P}(i_n | f)  = 1$. Therefore, in order to compute an estimate of the full joint PMF, we propose solving the following optimization problem
\begin{equation}
\begin{aligned}
&  \min_{ \mathbf{A}_1,\ldots,\mathbf{A}_N,\boldsymbol{\lambda} } \sum_{i,j,k} && \frac{1}{2} \left \| \underline{\mathbf{X}}_{ijk} - [\![ \boldsymbol{\lambda},\mathbf{A}_i,\mathbf{A}_j,\mathbf{A}_k ]\!] \right \|_F^2 \\
& \text{subject to}
& & \boldsymbol{\lambda}\geq \mathbf{0}, \\
&&& {\mathbf{1}}^T\boldsymbol{\lambda}=1, \\
&&& \mathbf{A}_n \geq \mathbf{0}, \; n=1\ldots N, \\
&&& {\mathbf{1}}^T \mathbf{A}_n = \mathbf{1}^T, \; n=1\ldots N,  \\
\end{aligned}
\label{eq:coupled_tensor}
\end{equation}
where $\mathbf{A}_n\in \mathbb{R}_+^{I_n \times F}$,  $n=1\ldots N$, $\boldsymbol{\lambda} \in \mathbb{R}_+^F$. The optimization problem in~\eqref{eq:coupled_tensor} is an instance of coupled tensor factorization. Coupled  tensor/matrix factorization has attracted a lot of interest lately, especially in data mining as a way of combining various datasets that share dimensions and corresponding matrix factors~\cite{BeTaKu2014,AcKoDu2011}. Notice that in the case where we have estimates of pair-wise marginals, the optimization problem in~\eqref{eq:coupled_tensor} corresponds to coupled matrix factorization.

An important question that arises is whether the parameters of the model in Equation~\eqref{eq:pmf_latent_var} are identifiable from the lower-order marginals. We know that this is {\em not} the case when only first-order marginals are given, unless the random variables are independent. But what about the case where third-order marginals are given? The answer in this case is affirmative. In fact is is possible to derive combinatorially-many identifiability results here, so we restrict ourselves to two illustrative ones. Uniqueness conditions for coupled CPD of third-order tensors {\em with one common factor} have been provided in~\cite{SoDe2015}, which showed that the coupling between several CPDs can in fact enhance identifiability relative to considering individual CPDs. The result in~\cite{SoDe2015} can be applied to derive identifiability conditions in our context. Another possibility is outlined next. Consider the third-order marginals for random variables (RVs) 1, 2, and a third RV. Using the mode-$1$ unfolding and stacking the marginal distributions
\begin{equation}
\begin{bmatrix}
\mathbf{X}^{(1)}_{123} \\
\mathbf{X}^{(1)}_{124} \\
\vdots				   \\
\mathbf{X}^{(1)}_{12N}
\end{bmatrix} =
\begin{bmatrix}
(\mathbf{A}_3 \odot \mathbf{A}_2) \mathcal{D}(\boldsymbol{\lambda})  \mathbf{A}_1^T \\
(\mathbf{A}_4 \odot \mathbf{A}_2)  \mathcal{D}(\boldsymbol{\lambda}) \mathbf{A}_1^T\\
\vdots				   \\
(\mathbf{A}_N \odot \mathbf{A}_2)  \mathcal{D}(\boldsymbol{\lambda}) \mathbf{A}_1^T
\end{bmatrix} =
\left(
\begin{bmatrix}
\mathbf{A}_3   \\
\mathbf{A}_4   \\
\vdots			\\	
\mathbf{A}_N
\end{bmatrix} \odot \tilde{\mathbf{A}}_2
\right)  \mathbf{A}_1^T,
\label{eq:coupled_cpd}
\end{equation}
where we have absorbed the scaling in $\tilde{\mathbf{A}}_2$. Let $\mathbf{M}_2(\mathbf{A})$ denote the $\binom{I}{2} \times \binom{J}{2}$ compound matrix \cite{SiDeFu2016} of $\mathbf{A} \in \mathbb{R}^{I \times J}$. If rank($\begin{bmatrix} \mathbf{A}_3^T \mathbf{A}_4^T  \cdots \mathbf{A}_N^T  \end{bmatrix}^T = F$) and rank($\mathbf{M}_2(\mathbf{A}_1) \odot \mathbf{M}_2(\tilde{\mathbf{A}}_2)) = \binom{F}{2}$ then the rank of the tensor is $F$ and the decomposition is essentially unique -- cf. Theorem 6  in~\cite{SiDeFu2016}. If $\prod_{n=3}^N I_n \geq F$, $\min(I_1,I_2)\geq 3$, and $(I_1-1)(I_2-1) \geq F$, then the rank of the tensor is $F$ and the decomposition is unique almost surely -- cf. Theorem 7 in~\cite{SiDeFu2016}. We have simply scratched the surface here; there are many more possibilities for proving identifiability under further relaxed conditions. For example, further exploiting the coupling between the factors by also accounting for the marginals that depend on  $\mathbf{A}_1,\mathbf{A}_3$, regarding $\mathbf{A}_1$ as the common factor of two CPDs.

\section{Alternating Optimization Based on ADMM.}
Alternating Optimization (AO) is one of the most commonly used methods for computing a constrained CPD model. In order to solve the optimization problem in~\eqref{eq:coupled_tensor} we develop an AO algorithm in which we cyclically update variables $\mathbf{A}_n, n=1,\ldots,N$ and $\boldsymbol{\lambda}$ while fixing the remaining variables at their last updated values. Problem~\eqref{eq:coupled_tensor} is non-convex but it becomes convex with respect to each variable if we fix the remaining ones. Assume that we fix estimates $\boldsymbol{\lambda},\mathbf{A}_n$, $n=1,\ldots,i-1,i+1,\ldots,N$. Then, the optimization problem with respect to $\mathbf{A}_i$ becomes
\begin{equation}
\begin{aligned}
&  \min_{ \mathbf{A}_i} \sum_{\substack{ j \\ j\neq i}} \sum_{ \substack{k \\ k \neq j \\ k \neq i }} && \frac{1}{2} \left \| \underline{\mathbf{X}}_{ijk} - [\![ \boldsymbol{\lambda},\mathbf{A}_i,\mathbf{A}_j,\mathbf{A}_k ]\!] \right \|_F^2 \\
& \text{subject to} & & \mathbf{A}_i \geq \mathbf{0}, \;  \\
&&& {\mathbf{1}}^T \mathbf{A}_i = \mathbf{1}^T.   \\
\end{aligned}
\end{equation}
Note that we have dropped the terms that do not depend on $\mathbf{A}_i$. The number of marginals that depend on the $i$-th variable is $\binom{N-1}{2}$. By using the mode-1 unfolding of each tensor $\underline{\mathbf{X}}_{ijk}$, the problem can be equivalently written as
\begin{equation}
\begin{aligned}
&  \min_{ \mathbf{A}_i} \sum_{\substack{ j \\ j\neq i}} \sum_{  \substack{k \\ k \neq j \\ k \neq i }}  && \frac{1}{2} \left \| \mathbf{X}_{ijk}^{(1)} - (\mathbf{A}_k \odot \mathbf{A}_j) \mathcal{D}(\boldsymbol{\lambda}) \mathbf{A}_i^T \right \|_F^2 \\
& \text{subject to} & & \mathbf{A}_i \geq \mathbf{0}, \;  \\
&&& {\mathbf{1}}^T \mathbf{A}_i = \mathbf{1}^T, \\
\end{aligned}
\label{eq:coupled_tensor_sub1}
\end{equation}
which is a least-squares problem with respect to matrix $\mathbf{A}_i$ under probability simplex constraints on its columns. Similar expressions can be derived for each factor $\mathbf{A}_n$ due to symmetry. Finally, in order to update $\boldsymbol{\lambda}$ we solve the following optimization problem
\begin{equation}
\begin{aligned}
&  \min_{ \boldsymbol{\lambda} } \sum_{i} \sum_{\substack{ j \\ j\neq i}} \sum_{ \substack{k \\ k \neq j \\ k \neq i }}  && \frac{1}{2} \left \| \text{vec}(\underline{\mathbf{X}}_{ijk}) - (\mathbf{A}_k \odot \mathbf{A}_j \odot \mathbf{A}_i) \boldsymbol{\lambda}  \right \|_2^2 \\
& \text{subject to} & & \boldsymbol{\lambda} \geq \mathbf{0}, \;  \\
&&& {\mathbf{1}}^T \boldsymbol{\lambda} = 1.  \\
\end{aligned}
\label{eq:coupled_tensor_sub2}
\end{equation}
We use the alternating direction method of multipliers (ADMM) for solving problems~\eqref{eq:coupled_tensor_sub1},~\eqref{eq:coupled_tensor_sub2}. Let $\mathcal{S} = \{ \mathbf{A}  \mid \mathbf{A}\geq 0, \mathbf{1}^T\mathbf{A} = \mathbf{1}^T\}$ be the convex set that represents the probability simplex constraints on the factors and define
\begin{equation}
r(\mathbf{A}) = \begin{cases}
0, \;\;\;  \mathbf{A} \in \mathcal{S}\\
\infty,  \; \mathbf{A} \notin \mathcal{S}
\end{cases},
\end{equation}
which is the indicator function of set $\mathcal{S}$.
We reformulate Problem~\eqref{eq:coupled_tensor_sub1} and write it as
\begin{equation}
\begin{aligned}
& \min_{ \mathbf{A}_i, \hat{\mathbf{A}} } \sum_{ \substack{ j\\ j \neq i}} \sum_{ \substack{k   \\ k \neq j \\ k \neq i }} && \frac{1}{2} \left \| {\mathbf{X}}_{ijk}^{(1)} - (\mathbf{A}_k \odot \mathbf{A}_j) \mathcal{D}(\boldsymbol{\lambda}) \hat{\mathbf{A}} \right \|_F^2  + r(\mathbf{A}_i) \\
& \text{subject to} & &  \mathbf{A}_i = \hat{\mathbf{A}}^T. \\
\end{aligned}
\label{eq:coupled_tensor_admm}
\end{equation}
It is easy to adopt the ADMM algorithm~\cite{Boyd10} and derive the following updates
\begin{equation}
\begin{aligned}
\hat{\mathbf{A}}  & =  (\mathbf{G}_i + \rho \mathbf{I})^{-1} (\mathbf{V}_i  + \rho ( \mathbf{A}_i + \mathbf{U}_i)^T), \\
\mathbf{A}_i 	 & =  \argmin_{\mathbf{A}_i} r(\mathbf{A}_i) + \frac{\rho}{2} \| \mathbf{A}_i - \hat{\mathbf{A}}^T + \mathbf{U}_i \|_F^2, \\
\mathbf{U}_i	 	 & =  \mathbf{U}_i + \mathbf{A}_i - \hat{\mathbf{A}}^T,
\end{aligned}
\end{equation}
where
\begin{equation}
\mathbf{G}_i = \sum_{ \substack{ j\\ j \neq i}} \sum_{ \substack{k   \\ k \neq j \\ k \neq i }}  \mathcal{D}(\boldsymbol{\lambda}) \mathbf{Q}_{kj}^T\mathbf{Q}_{kj}  \mathcal{D}(\boldsymbol{\lambda}),
\end{equation}
\begin{equation}
\mathbf{V}_i = \sum_{ \substack{ j\\ j \neq i}} \sum_{ \substack{k   \\ k \neq j \\ k \neq i }} \mathcal{D}(\boldsymbol{\lambda}) \mathbf{Q}_{kj}^T {\mathbf{X}}_{ijk}^{(1)},
\end{equation}
\begin{equation}
\mathbf{Q}_{kj} = \mathbf{A}_k \odot \mathbf{A}_j.
\end{equation}

The update of $\mathbf{A}_i$ is called proximity operator of the function $\frac{1}{\rho}r(\cdot)$ which is the indicator function of a convex set, and thus a projection operator. To project onto the probability simplex we use a simple $O(n\log n)$ complexity algorithm~\cite{WaCa13}. Note that in order to efficiently compute matrix $\mathbf{G}_i$  we use a property of the Khatri-Rao product.
\begin{equation}
\mathbf{G}_i = \sum_{ \substack{ j\\ j \neq i}} \sum_{ \substack{k   \\ k \neq j \\ k \neq i }} \mathcal{D}(\boldsymbol{\lambda}) \left[ (\mathbf{A}_k^T\mathbf{A}_k )\circledast(\mathbf{A}_j^T\mathbf{A}_j) \right]  \mathcal{D}(\boldsymbol{\lambda}).
\end{equation}
Very efficient algorithms also exist for the computation of matrix $\mathbf{V}_i$ which is a sum of Matricized Tensor Times Khatri-Rao Product (MTTKRP) terms~\cite{BaKo2008,SmiRaSi2015}.
Similarly, we can derive updates for $\boldsymbol{\lambda}$.
\begin{equation}
\begin{aligned}
\hat{\boldsymbol{\lambda}} & =  (\mathbf{G} + \rho \mathbf{I})^{-1} (\mathbf{V}  + \rho ( \boldsymbol{\lambda} + \mathbf{u})), \\
\boldsymbol{\lambda} & =  \argmin_{\boldsymbol{\lambda}} r(\boldsymbol{\lambda}) + \frac{\rho}{2} \| \boldsymbol{\lambda} - \hat{\boldsymbol{\lambda}} + \mathbf{u} \|_F^2, \\
\mathbf{u} & =  \mathbf{u} + \boldsymbol{\lambda} - \hat{\boldsymbol{\lambda}}.
\end{aligned}
\end{equation}
In this case we need to compute matrices
\begin{equation}
\mathbf{G} =  \sum_i \sum_{ \substack{ j\\ j \neq i}} \sum_{ \substack{k   \\ k \neq j \\ k \neq i }}  \mathbf{Q}_{kji}^T\mathbf{Q}_{kji},
\end{equation}
\begin{equation}
\mathbf{V} = \sum_i  \sum_{ \substack{ j\\ j \neq i}} \sum_{ \substack{k   \\ k \neq j \\ k \neq i }} \mathbf{Q}_{kji}^T \text{vec}(\underline{\mathbf{X}}_{ijk}),
\end{equation}
\begin{equation}
\mathbf{Q}_{kji} = \mathbf{A}_k \odot \mathbf{A}_j \odot \mathbf{A}_i.
\end{equation}
Matrix $\mathbf{G}$ can be computed using the property of the Khatri-Rao product. Matrix $\mathbf{V}$ can be efficiently computed without explicitly forming the Khatri-Rao products and also by exploiting sparsity in $\text{vec}(\underline{\mathbf{X}}_{ijk})$.  We run the ADMM algorithm for each subproblem until the primal and dual residuals are below a certain threshold~\cite{Boyd10} or a maximum number of iterations has been reached.

\section{Numerical Results}
In this section, we evaluate our method on both synthetic and real datasets.

\subsection{Synthetic Dataset}
 We generate a low-rank five-way tensor with factor matrices $\mathbf{A}_n \in \mathbb{R}_+^{I_n \times F}$, $I_n = 10$, $n=1,\ldots,5$, $\boldsymbol{\lambda} \in \mathbb{R}_+^F$ and $F \in \{5,10,15\}$. The elements of each factor and the vector $\boldsymbol{\lambda}$  are drawn from an i.i.d uniform distribution between zero and one and are normalized so that the tensor elements  sum up to one. We simulate scenarios where we are given different noiseless marginal distributions of the PMF as input, that is we observe only projections of the original tensor. The different types of input are pair-wise marginals, marginals of triples and quadruples of the random variables. We run 20 Monte Carlo simulations with randomly generated tensors and compute the mean relative error of the factors as well as the mean relative error of the recovered tensor which are defined as follows
\begin{equation}
\text{MRE} _{\text{fact}}= \frac{1}{NK} \sum_{k=1}^{K}  \sum_{n=1}^{N} \frac{ \| {\mathbf{A}}_n^k - \Pi^k \hat{{\mathbf{A}}}_n^k \|_F}{ \| {\mathbf{A}}_n \|_F},
\end{equation}
\begin{table}[!t]
\begin{center}
\begin{tabular}{c  l  c  c}
\hline
 Rank   &  	            & Rel. Fact. Error & Rel. Ten. Error\\
\hline
  		&  Pairs        			&   	$0.235$			    				&   $0.124$            \\
$F = 5$ 		&  Triples     			&   	$1.24 \times 10^{-6}$				& 	$2.80 \times 10^{-7}$		       \\
	    &  Quadruples   			&   	$8.64 \times 10^{-11}$				& 	$1.53 \times 10^{-11}$		       \\

\hline
  		&  Pairs        			&   	$0.412$			    &   $ 0.176$            \\
$F = 10$		&  Triples     			&   	$6.91 \times 10^{-5}$				& 	$1.36 \times 10^{-5}$		       \\
	    &  Quadruples   			&   	$2.17 \times 10^{-9}$				& 	$3.37 \times 10^{-10}$		       \\
\hline
  		&  Pairs        			&   	$0.433$			    &   $ 0.194$            \\
$F = 15$			&  Triples   			&   	$8.56 \times 10^{-4}$							& 	$1.47 \times 10^{-4}$		       \\
	    &  Quadruples   			&   	$8.95 \times 10^{-7}$				& 	$3.63 \times 10^{-8}$		       \\
\hline
\end{tabular}
\end{center}
\caption{Relative factor and tensor error for different choices of rank in noiseless data.}
\label{table:CPD_noiseless}
\end{table}
\begin{equation}
\text{MRE}_{\text{ten}} = \frac{1}{K} \sum_{n=1}^{K} \frac{ \| \underline{\mathbf{X}}^k - \hat{\underline{\mathbf{X}}}^k \|_F}{ \| \underline{\mathbf{X}}^k \|_F},
\end{equation}
where ${\mathbf{A}}_n^k$ is the $k$-th realization of the $n$-th factor, $\underline{\mathbf{X}}^k$ is the $k$-th tensor realization, $\Pi^k$ is a permutation matrix to fix the inherent permutation ambiguity and $\hat{\underline{\mathbf{X}}}^k$, $\hat{{\mathbf{A}}}_n^k$ are the estimated tensor and the corresponding factors.
We run the alternating optimization algorithm based on ADMM  until the maximum number of iterations is met which was set to $1500$. Table~\ref{table:CPD_noiseless} shows the mean relative factor and tensor error for the different types of input and different choices of rank. We observe that using triples or quadruples of random variables we are able to identify the true model parameters. On the other hand, using every combination of pair-wise marginals we observe that we are not able to identify the true model parameters.

We repeated the above experiments using a slightly perturbed five-way tensor. After generating the low-rank five-way tensor, we add white Gaussian noise with standard deviation $\sigma = 10^{-6}$. The low-rank tensor is then projected onto the probability simplex. Table~\ref{table:CPD_noisy} shows the mean relative factor and tensor error for the noisy data. Similarly to the noiseless case, we observe that using triples or quadruples of random variables we are able to achieve low relative tensor and factor errors.

\subsection{Real Datasets}
Next, we evaluate the performance of our method in real datasets. More specifically, we test our method in the task of rating prediction. We compute a low-rank CPD model of a joint PMF by using lower-order marginals of pairs, triples and quadruples of variables that we estimate using a training set. Then, we use the estimated PMF in order to compute the expected value of  users' ratings that we do not observe given the ones we observe. As a baseline algorithm we use the Biased Matrix Factorization (BMF) method~\cite{Koren2009}. We test our method using two collaborative filtering datasets, Movielens and Jester.

MovieLens~\cite{HaMaKo2015} is a collaborative filtering dataset that contains 5-star movie ratings with 0.5 star increments. In order to test our algorithm we select three different subsets of the full dataset. Three different categories (action, animation and romance) are selected first. From each category we extract a small and relatively dense submatrix by keeping the 10 most rated movies.

\begin{table}[!t]
\begin{center}
\begin{tabular}{c  l  c  c}
\hline
 Rank   &  	            & Rel. Fact. Error & Rel. Ten. Error\\
\hline
  		&  Pairs        			&   	$0.305$			    				&   $0.17$       \\
$F = 5$ 		&  Triples     		&   	$4.5 \times 10^{-3}$				& 	$4.4 \times 10^{-3}$		 \\
	    &  Quadruples   			&   	$4.1 \times 10^{-3}$				& 	$4  \times 10^{-3}$		     \\

\hline
  		&  Pairs        			&   	$0.41$			    &   $ 0.181$            \\
$F = 10$		&  Triples     			&   	$10.3 \times 10^{-3}$				& 	$6.7 \times 10^{-3}$	 \\
	    &  Quadruples   			&   	$9.2 \times 10^{-3}$				& 	$6.1 \times 10^{-3}$		 \\
\hline
  		&  Pairs        			&   	$0.428$			    &   $ 0.19$            \\
$F = 15$			&  Triples   			&   	$16.2 \times 10^{-3}$							& 	$8.4 \times 10^{-3}$		       \\
	    &  Quadruples   			&   	$14.1 \times 10^{-3}$				& 	$7.7 \times 10^{-3}$		       \\
\hline
\end{tabular}
\end{center}
\caption{Relative factor and tensor error for different choices of rank in noisy data with $\sigma = 10^{-6}$.}
\label{table:CPD_noisy}
\end{table}
Jester~\cite{GoRo2001} is a collaborative filtering dataset that contains  continuous ratings (-$10.00$ to +$10.00$) of 100 jokes. Again, we extract a dense submatrix corresponding to $10$ jokes that have been rated from almost all users. The resulting dataset is processed such that the ratings correspond to integers ($1$ to $21$) by rounding each continuous rating.

For each dataset we randomly hide $20\%$ ratings that we use as a test set, $10\%$ ratings that we use as a validation set and the remaining dataset is used as a training set. We run $20$ Monte Carlo simulations using the alternating optimization ADMM  and BMF algorithms until no improvement is observed in the cost function. At each iteration we calculate the Root Mean Squared Error (RMSE) based on the validation set and after the algorithm converges we return the model that reports the best RMSE on the validation set.

\begin{table*}[!ht]
\begin{center}
\begin{tabular}{l|  c  c | c c | c c | c c}
&  \multicolumn{2}{c}{\textbf{MovieLens Dataset 1}}            &  \multicolumn{2}{c}{\textbf{MovieLens Dataset 2}}     &  \multicolumn{2}{c}{\textbf{MovieLens Dataset 3}} &  \multicolumn{2}{c}{\textbf{Jester Dataset}}\\
\hline
 Method &  RMSE       & 	MAE     &  RMSE      			& MAE  &  RMSE      & MAE  &  RMSE      & MAE 	\\
\hline
 CP (Pairs)       & $0.8095$      	    & 	$0.6134$   			&  $0.7637$ 	& $0.5811$  &  $0.9038$          & $0.7028$  & $4.8585$  & $4.0052$ \\
 CP (Triples)     & $0.7903$     		& 	$0.6003$   			&  $0.7443$     & $0.5655$  &  $0.8955$          & $0.6947$  	& $4.7931$  & $3.9270$\\
 CP (Quadruples)   & $\mathbf{0.7874}$    & 	$\mathbf{0.5994}$   &  $\mathbf{0.7419}$ & $\mathbf{0.5624}$  &  $\mathbf{0.8912}$       & $\mathbf{0.6916}$  & $\mathbf{4.7797}$  & $\mathbf{3.9204}$	 \\
\hline
 Global Average   & $0.9368$ & 	$0.7157$     &  $0.8924$      & $0.7026$ &  $1.0102$   & $0.8175$  &  $ 5.1975$   & $4.4201$
	   \\
 User  Average  	  & $0.9388$ & 	$0.6979$  	    &  $0.8008$      & $0.5787$ &  $1.0693$   & $0.8106$    &   $5.0287$  & $ 4.0377$
	 \\
 Item Average  	  & $0.8888$ & 	$0.6863$  	    &  $0.8864$      & $0.6930 $ &  $0.9549$      & $0.7516$  & $5.1063$ & $4.3148$	 \\
 BMF  	          & $0.8161$ & 	$0.6367$   	&  $0.7443$ & $0.5760$ &  $0.9207$   & $0.7293$        & $4.8168$  &$4.0088$     	 \\
\hline
\end{tabular}
\end{center}
\caption{RMSE and MAE of different algorithms on MovieLens (Ratings are in the range [0.5-5] )  and Jester (Ratings are in the range [1-21]) datasets.}
\label{table:real_dataset}
\end{table*}
\begin{figure*}[!ht]
\begin{subfigure}{0.24\textwidth}
\includegraphics[width=1\textwidth]{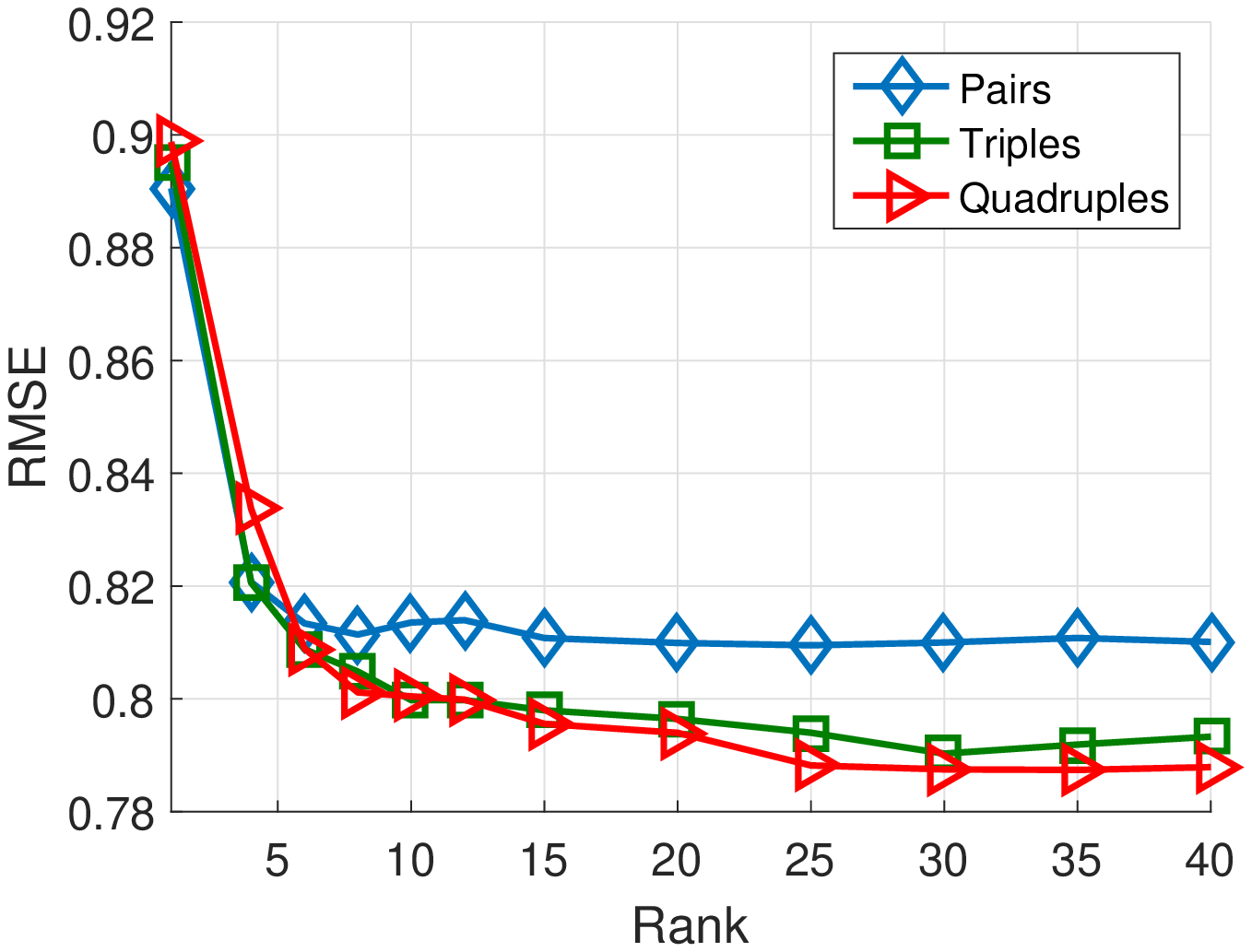}
\caption{MovieLens dataset 1.}
\end{subfigure}\hspace{\fill}
\begin{subfigure}{0.24\textwidth}
\includegraphics[width=1\textwidth]{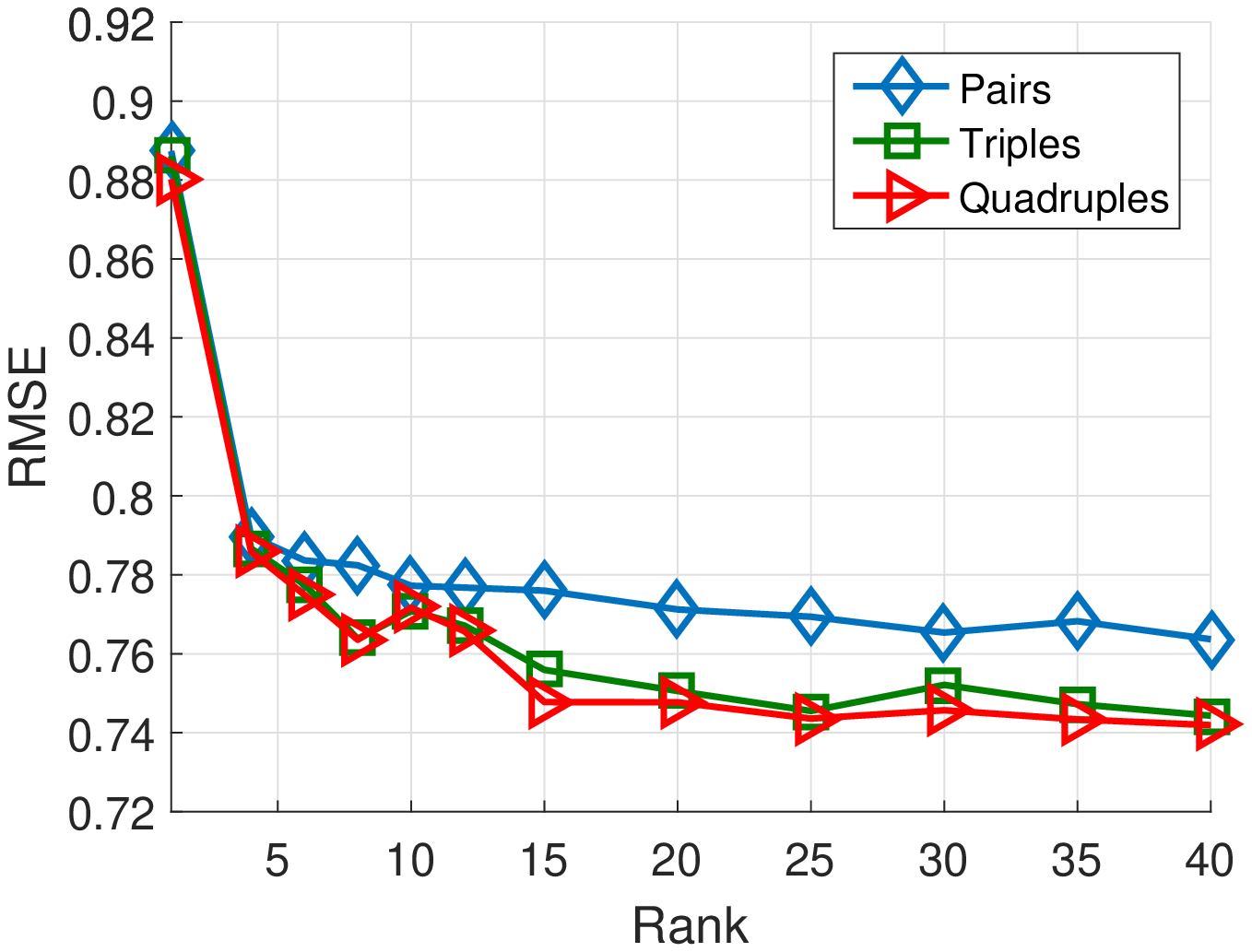}
\caption{MovieLens dataset 2.}
\end{subfigure}\hspace{\fill}
\begin{subfigure}{0.24\textwidth}
\includegraphics[width=1\textwidth]{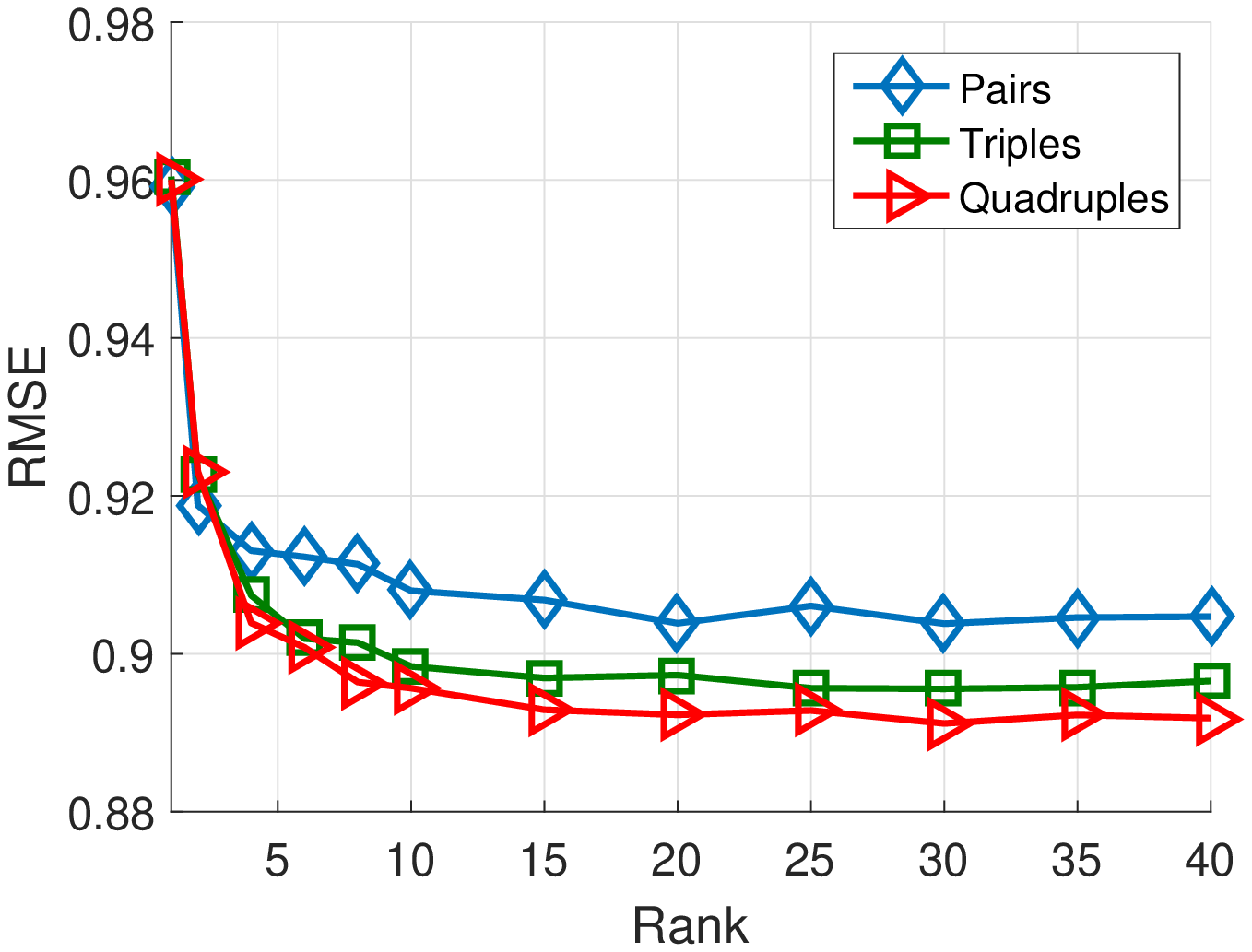}
\caption{MovieLens dataset 3.}
\end{subfigure}\hspace{\fill}
\begin{subfigure}{0.24\textwidth}
\includegraphics[width=1\textwidth]{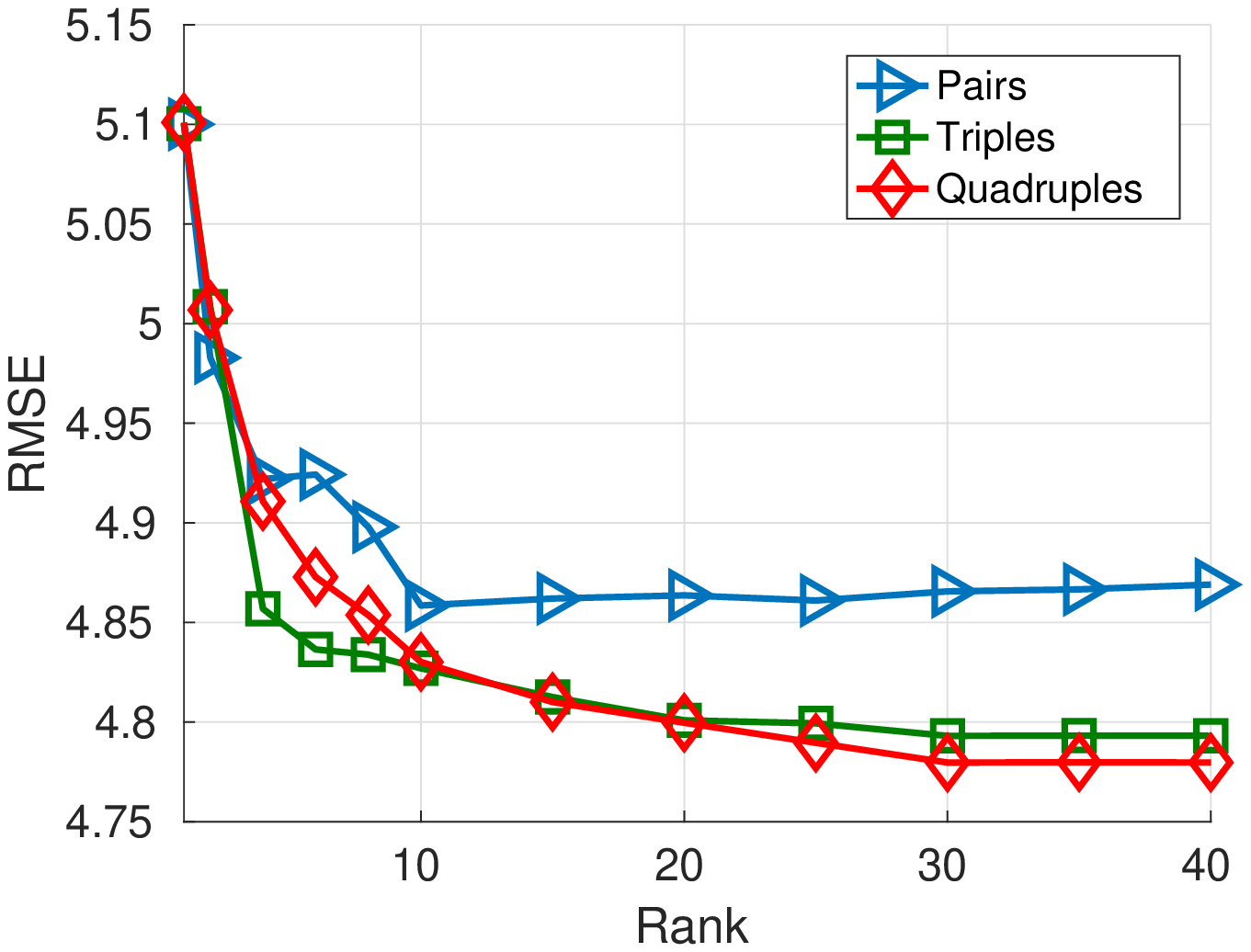}
\caption{Jester dataset.}
\end{subfigure}\hspace{\fill}
\caption{RMSE as a function of rank.}
\label{fig:rmsevsrank}
\end{figure*}
Table~\ref{table:real_dataset} shows the performance of the two algorithms in terms of the RMSE and Mean Absolute Error (MAE). For our method, we tested different values for the rank parameter and report the model that performed the best on the test set. Similarly, for BMF we tested different values for both the rank and the regularization parameter and reported the model that performed the best on the test set. In addition, we present the RMSE and MAE obtained when we use as our prediction the global average of the ratings, the user average and the item average. We observe that the performance of our method is better than the baselines when we use information corresponding to marginals of triples and quadruples of the random variables.

Figure~\ref{fig:rmsevsrank} shows the performance of our method for different values of rank. We observe that the behavior of the algorithm for the different datasets is similar. As rank increases the RMSE drops until it reaches a plateau.

\section{Conclusion}
In this work, we proposed a method based on tensor decomposition for computing a parsimonious model of a joint PMF using lower-order marginals. We formulated the problem as coupled tensor factorization and described an algorithmic approach to solve it. When the joint PMF admits a low-rank decomposition we showed that it is possible to recover the true factors using synthetic data. Finally, we showed some preliminary results in collaborative filtering datasets for rating prediction.

\bibliographystyle{IEEEtran}
\bibliography{lib}
\end{document}